\documentclass[lettersize,journal]{IEEEtran}
\usepackage{amsmath,amsfonts}
\usepackage{algorithmic}
\usepackage{algorithm}
\usepackage{array}
\usepackage[caption=false,font=normalsize,labelfont=sf,textfont=sf]{subfig}
\usepackage{textcomp}
\usepackage{stfloats}
\usepackage{url}
\usepackage{verbatim}
\usepackage{graphicx}
\usepackage{cite}
\usepackage{hyperref}
\hypersetup{colorlinks=true, urlcolor=magenta}
\begin{document}

\title{Towards 3D Face Reconstruction in Perspective Projection: Estimating 6DoF Face Pose from Monocular Image}

\author{Yueying Kao, Bowen Pan, Miao Xu, Jiangjing Lyu, Xiangyu Zhu, Yuanzhang Chang,  Xiaobo Li, and  Zhen Lei

\thanks{Yueying Kao, Bowen Pan, Jiangjing Lyu, Yuanzhang Chang and Xiaobo Li are with Alibaba Group. Email: \{kaoyueying.kyy, bowen.pbw, jiangjing.ljj, yuanzhang.cyz, xiaobo.lixb\}@alibaba-inc.com. Xiangyu Zhu, Zhen Lei, and Miao Xu are with Center for Biometrics and Security Research \& State Key Laboratory of Multimodal Artificial Intelligence Systems, Institute of Automation, Chinese Academy of Sciences (CASIA), Beijing 100190, China, and also with the School of Artificial Intelligence, University of Chinese Academy of Sciences (UCAS), Beijing 100049, China. Zhen Lei is also with the Centre for Artificial Intelligence and Robotics, Hong Kong Institute of Science \& Innovation, Chinese Academy of Sciences, Hong Kong, China. Email: \{xiangyu.zhu, zlei\}@nlpr.ia.ac.cn, xumiao2021@ia.ac.cn. 

This work is supported in part by National Key Research \& Development Program (No. 2020AAA0140000), Alibaba Group through Alibaba Innovative Research Program,  Chinese National Natural Science Foundation Projects \#62176256, \#62276254, and the InnoHK program.

Xiangyu Zhu is the corresponding author. }

}


\maketitle

\begin{abstract}
In 3D face reconstruction, orthogonal projection has been widely employed to substitute perspective projection to simplify the fitting process. This approximation performs well when the distance between camera and face is far enough. However, in some scenarios that the face is very close to camera or moving along the camera axis, the methods suffer from the inaccurate reconstruction and unstable temporal fitting due to the distortion under the perspective projection. In this paper, we aim to address the problem of single-image 3D face reconstruction under perspective projection. Specifically, a deep neural network, Perspective Network (PerspNet), is proposed to simultaneously reconstruct 3D face shape in canonical space and learn the correspondence between 2D pixels and  3D points, by which the 6DoF (6 Degrees of Freedom) face pose can be estimated to represent perspective projection. Besides, we contribute a large ARKitFace dataset to enable the training and evaluation of 3D face reconstruction solutions under the scenarios of perspective projection, which has 902,724 2D facial images with ground-truth 3D face mesh and annotated 6DoF pose parameters. Experimental results show that our approach outperforms current state-of-the-art methods by a significant margin. The code and data are available at \href{https://github.com/cbsropenproject/6dof\_face}{https://github.com/cbsropenproject/6dof\_face}.
\end{abstract}


\section{Introduction}
\label{sec:intro}

\begin{figure}
\centering
\includegraphics[height=6cm]{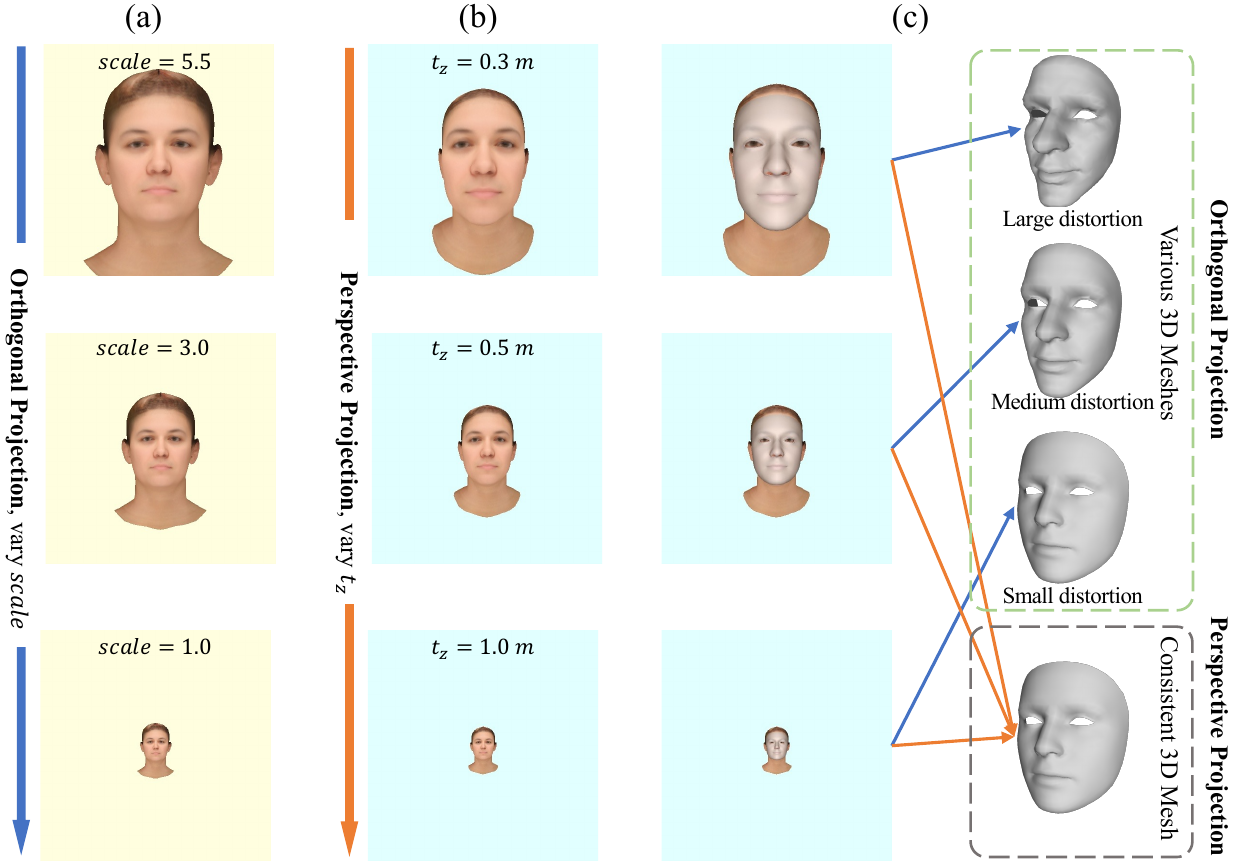}
\caption{Orthogonal projection vs. Perspective projection. (a) and (b) show rendered images with a same 3D face by changing the pose parameters that represent the size in the two projections, respectively. The  rendered 2D faces are only zoomed in and out with scale variation in orthogonal projection, while in perspective projection there exists obvious distortion by changing $t_z$, especially in very near distance. (c) shows 3D face reconstruction under the two projections respectively. Orthogonal projection based methods explain the distortion by the shape changes, while perspective projection methods provide the same shape and explain it by different pose parameters.} 
\label{fig1}
\end{figure}

6DoF pose estimation with/without 3D face reconstruction~\cite{yang2019fsa,albiero2021img2pose,romdhani2005estimating,feng2018joint,bai2021riggable,feng2021learning,zhu2020beyond,tewari2017mofa}  have draw much attention recently in computer vision and computer graphics communities, due to the increasing demand from many applications, such as virtual glasses try-on and make-up in AR, video editing and animation. Most of 3D face reconstruction methods~\cite{zhu2017face,feng2018joint,bai2021riggable,feng2021learning,guo2020towards} employ the orthogonal projection to approximate the real-world perspective projection, which works well when the size of face is small compared to the distance from the camera (roughly 1/20 of the camera distance)~\cite{geometry_persp,forsyth2011computer,fried2016perspective,zhao2019learning}. However, the scenarios of face capture become more complicated with the popular of selfies, virtual glasses try-on and makeup, etc. When the subject is very close to the camera, the rendered 2D faces in orthogonal projection are only zoomed in, while in perspective projection there exists obvious distortion in the rendered 2D faces, especially in the very close distance, as shown in Fig.~\ref{fig1}(a) and (b). Under the approximation of orthogonal projection, the distortion by perspective projection is explained by the shape changes, leading to two significant problems: 1) This distortion is not modeled in the shape models, and the distorted faces are often outside of the shape space, which leads to the unstable temporal fitting.  2) When the subject moves along the camera axis $t_z$, the orthogonal projection based methods predict different face shapes in different distances due to the distortion, while perspective projection methods provide the consistent shape across frames since it explains the distortions with different 6DoF pose, as shown in Fig.~\ref{fig1}(c). Even introducing 6DoF pose apparently improves the accuracy and robustness of 3D face reconstruction, which benefits many VR/AR applications, 6DoF pose estimation for faces is still a challenging problem. Since we should capture the distortion by perspective projection from the large variations of face appearances under complicated environment. Besides, 6DoF pose estimation in other objects~\cite{peng2019pvnet,he2020pvn3d,tian2020shape} always assumes a pre-defined 3D shape but we only have a face image as input. 

To address the problem, we propose a new approach to recover 3D facial geometry under perspective projection by estimating the 6DoF pose, i.e. 3D orientation and translation, simultaneously from a single RGB image (see Section~\ref{sec:method}). Specifically, 6DoF estimation depends on 3 sub-tasks: 1) Reconstructing 3D face shape in canonical space, 2) Estimating pixel-wise 2D-3D correspondence between the canonical 3D space and image space and 3) Calculating 6DoF pose parameters by the Perspective-n-Point (PnP) algorithm~\cite{lepetit2009epnp}. 
In this paper, we propose a deep learning based Perspective Network (PerspNet) to achieve the goal in one propagation. For the 3D face shape reconstruction, an encoder-decoder architecture is designed to regress a UV position map~\cite{feng2018joint} from image pixels, which records the specific position information of a 3D face in canonical space. For the 2D-3D correspondence, we construct a sophisticated point description features including aligned-pixel features, 3D features and 2D position features, and map it to a 2D-3D correspondence matrix where each row records the corresponding 3D points in canonical space of one pixel on the input image. With 3D shape and correspondence matrix, 6DoF face pose can be robustly estimated by performing PnP. 

\begin{figure*}[t]
\centering
\includegraphics[height=3.8cm]{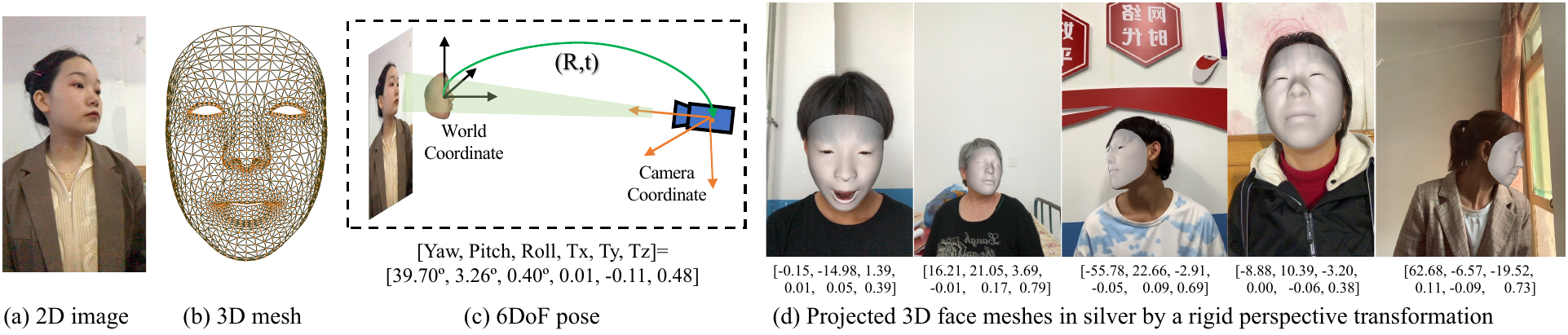}
\caption{Some examples of ARKitFace dataset. Each sample contains (a) 2D image, (b) 3D mesh and (c) 6DoF pose annotation. (d) shows more examples.}
\label{fig:arkit_face_example}
\end{figure*}

To realize 6DoF pose estimation of 3D faces, we need the face images, 3D shapes and annotated camera parameters for each training sample. However, current datasets cannot satisfy our requirement. Most of existing data provide the low-quality 3D face by the optimization-based fitting algorithm in a weak supervised manner, such as AFLW2000-3D~\cite{zhu2017face}. Some datasets, such as BIWI~\cite{fanelli2013random}, NoW~\cite{sanyal2019learning}, FaceScape~\cite{yang2020facescape}, lack either exact 3D shape for each 2D image or pose variations. Therefore, we construct a real large 3D face dataset called ARKitFace dataset, with 902,724 2D facial images of 500 subjects taken in diverse of expressions, ages and poses. For each 2D facial image, ground-truth 3D face mesh and 6DoF pose annotations are provided, as shown in Fig.~\ref{fig:arkit_face_example}. Compared with other 3D face datasets~\cite{zhu2017face,sanyal2019learning,fanelli2013random,yang2020facescape}, the ARKitFace is a very large amount dataset for single-image 3D face reconstruction under perspective projection.

To summarize, the main contributions of this work are:
\begin{itemize}
\setlength{\itemsep}{0pt}
\setlength{\parsep}{0pt}
\setlength{\parskip}{0pt}
\item[-] We explore a new problem of 3D face reconstruction under perspective projection, which will provide the exact 3D face shape and location in 3D camera space. 
\item[-] We propose PerspNet to reconstruct 3D face shape and 2D-3D correspondence simultaneously, by which the 6DoF face pose can be estimated by PnP.
\item[-] To enable the training and evaluation of PerspNet, we collect ARKitFace dataset, a large-scale 3D dataset with ground-truth 3D face mesh and 6DoF pose annotations for each image. 
\item[-] Experimental results on ARKitFace dataset and a public BIWI dataset show that our approach outperforms the state-of-the-art methods. The code and all data will be released to public under the authorization of all the subjects.
\end{itemize}

\section{Related Work}
 
\subsection{3D Face Reconstruction}
3D face reconstruction from a single RGB image is essentially an ill-posed problem. Most methods tackle this problem by estimating the parameters of a statistical face model~\cite{guo2020towards,bai2021riggable,zhu2017face,feng2018joint,feng2021learning,romdhani2005estimating,zhu2020beyond,brunton2014multilinear,Liu_2017_ICCV,tran2018extreme,guo20213d,Chinaev2019,Tu21,lattas2020avatarme,lattas2021avatarme++,yang2021self,wang2022faceverse,danvevcek2022emoca}.
Although these methods achieve remarkable results, they suffer from the same fundamental limitation: an orthogonal or weak perspective camera model is utilized when reconstructing the 3D face shape. The deformation caused by perspective projection, especially in the near distance, has to be compensated by face shape. For MOFA~\cite{tewari2017mofa}, MobileFace~\cite{Chinaev2019} and Deng \emph{et al.} ~\cite{deng2019accurate}, they aim to use perspective projection for 3D face reconstruction. But they use the cropped face region for optimization. The perspective projection parameters should be converted from those of full image, which is proved by img2pose method~\cite{albiero2021img2pose}. In addition, Deng \emph{et al.}~\cite{deng2019accurate} only optimize the 6DoF pose parameters without ground truth pose supervision. Thus, we insist on following the rule of perspective projection and believe there is still substantial room for improvement on the task of 3D face reconstruction.

 Zhao \emph{et al.} \cite{zhao2019learning} and Fried \emph{et al.} \cite{fried2016perspective} also consider the distortion of perspective projection. But there are some differences with our work. 1) Task: We aim to reconstruct exact 3D face shape and estimate 6DoF pose from a single image, while the two works\cite{zhao2019learning,fried2016perspective} focus on manipulation of portrait photos considering perspective distortion.  Specifically, Zhao \emph{et al.} \cite{zhao2019learning} aim to remove perspective distortion from a near-range portrait. Fried \emph{et al.} \cite{fried2016perspective} aim to edit distortion of portrait given a distance. 2) Method: We propose PerspNet to reconstruct 3D face shape and 2D-3D correspondence simultaneously, by which the 6DoF face pose can be estimated by PnP. Zhao \emph{et al.} \cite{zhao2019learning} remove perspective distortion by a framework with the supervision of undistorted portraits. Then the predicted undistorted portraits can be used to reconstructed 3D face by a fitting a 3D face model. Fried \emph{et al.} \cite{fried2016perspective} propose an iterative fitting method for image editing. 3) Evaluation: Our evaluation is designed for 3D face reconstruction and head pose estimation with SOTA methods. The two works \cite{zhao2019learning,fried2016perspective} focus on image editing and only give some reconstructed 3D face for the undistorted portraits.
 
 In addition, Apple device, such as iPhone 11, with ARKit toolbox can use both RGB and depth cameras to implement this 3D face reconstruction and pose estimation. The main purpose of our method is to support AR applications with only monocular RGB inputs, such as AR interactive games on Android phones, which can create the illusion that the real face interacts with virtual objects, even though there is only a single RGB image. For example, users can place virtual glasses on their faces with an RGB camera. The virtual glasses look like real.

\subsection{Head/Face Pose Estimation}
Due to the prevalence of deep learning, great progresses have been achieved in head pose estimation. The main idea is to regress the euler angles of head pose directly based on deep CNNs. 
QuatNet~\cite{hsu2018quatnet} addresses the non-stationary property of head pose and proposes to train a quaternions regressor to avoid the ambiguity problem in euler angles.
FSA-Net~\cite{yang2019fsa} adopts a fine-grained structure mapping for spatially feature grouping to improve the performance of head pose estimation.
EVA-GCN~\cite{xin2021eva} views the head pose estimation as a graph regression problem, and leverages the Graph Convolutional Networks to model the complex nonlinear mappings between the graph typologies and the head pose angles. All of the above methods are proposed under the assumption of orthogonal projection, thus only 3DoF (3D rotation) is predicted and the error caused by perspective deformation can not be avoided.

Laterly, Chang \emph{et al.}~\cite{chang2019deep,chang17fpn} regress directly local 6DoF human faces from a cropped face image. The predicted 6DoF pose is evaluated on downstream face recognition and face alignment tasks, which shows effectiveness of this 6DoF face pose. Albiero \emph{et al.}~\cite{albiero2021img2pose} propose a Faster-RCNN~\cite{ren2015faster} like framework named img2pose to predict the full 6DoF of human faces in the setting of perspective projection. Firstly, the original full image is fed into a region
proposal network (RPN) module, and all faces in the image are detected. Then the local 6DoF of each detected face is predicted by the ROI head. Finally the local 6DoF is converted to global 6DoF using the information of facial bounding box and camera intrinsic matrix. Since img2pose ignores the influence of perspective deformation, the error of 6DoF will be amplified during local-to-global conversion. In addition, the 6DoF pose annotations in their training data are calculated with predicted five landmarks and a mean 3D mesh, not real 6DoF pose.

\subsection{3D Face Datasets}
Although a large number of facial images are available, the corresponding 3D annotations are expensive and difficult to obtain. For 3D face shape reconstruction,  several 3D datasets are built, such as Bosphorus~\cite{savran2008bosphorus}, BFM~\cite{paysan20093d}, FaceWarehouse~\cite{cao2013facewarehouse}, MICC~\cite{bagdanov2011florence}, 3DFAW~\cite{pillai20192nd}, BP4D~\cite{zhang2014bp4d}, NoW dataset~\cite{sanyal2019learning}. In these datasets, there are either limited data or no face pose annotation. 

To obtain head/face 6DoF pose, some datasets synthesize 3D ground-truth, i.e. the parameters of a statistical face model, by the optimization-based fitting algorithm in a weak supervised manner, such as 300W-LP and AFLW2000-3D~\cite{zhu2017face}. For AFLW2000-3D dataset, its 3D face is provided by 3DMM fitting in a weak supervised manner and its projection follows orthogonal projection, which may not be proper for us since we aim to estimate 6DoF (perspective projection). Thus we do not give the results of AFLW2000-3D. The synthesized 3D ground-truth is coarse because only the reconstruction error of 2D sparse facial landmarks is considered. Despite the expensiveness of 3D annotations, researchers have collected several 3D face datasets using professional imaging devices for the sake of high precision on the tasks of 3D face reconstruction and head pose estimation. For example, BIWI dataset~\cite{fanelli2013random} is captured by a Kinect sensor with global rotation and translation to the RGB camera. However, the number of individuals is limited, and facial images with only neutral expression are recorded. Most importantly, the size of BIWI dataset is too small for learning deep networks. FaceScape~\cite{yang2020facescape} is a large-scale dataset recorded by multi-view camera system in an extremely constrained environment. It contains sufficient individuals and multiple facial expressions. However, the head pose is fixed by limited number of camera locations and the lighting condition is constant. Models that are trained on such dataset do not have good generalization ability to the real scenario in the wild. Different from these datasets, we aim to collect a 3D dataset with 3D mesh and 6DoF pose estimation in different conditions, such as expression, age and 6D pose variations.

\section{Proposed Method}
\label{sec:method}
In this paper, we propose a novel framework for 3D face reconstruction under perspective projection from a single 2D face image. In previous single-image 3D face reconstruction method~\cite{zhu2017face,feng2018joint,bai2021riggable,feng2021learning,guo2020towards}, scaled orthographic projection camera model is adopted to project 3D face shape into image space. We denote the 3D face shape as $X \in \mathbb{R}^{3 \times n}$ representing $n$ 3D vertices (points) on the surface of the 3D shape in world coordinate system (canonical space). This projection process is usually formulated as 

\begin{equation}
V=s \Pi(X) + t_{2d},
  \label{eq:first}
\end{equation}
where $V \in \mathbb{R}^{2 \times n}$ denotes projected 2D coordinates in 2D image of $X$,  $\Pi \in \mathbb{R}^{2 \times 3}$ is the orthographic 3D-2D projection matrix, $s$ is isotropic scale and $t_{2d} \in \mathbb{R}^{2}$ denotes 2D translation.  Different from orthogonal projection, in perspective projection,  3D face shape $X$ is firstly transformed from the world coordinate system to the camera coordinate system by using 6DoF face pose $(R,t)$, 
\begin{equation}
X_c=K(RX+t),
  \label{eq:first}
\end{equation}
with known intrinsic camera parameters $K$, where $R$, $t$ and $X_c$ represent the 3D rotation $ R \in SO(3)$,  the 3D translation $t \in \mathbb{R}^3$, and 3D face vertices in camera coordinate system. Then the $X_c$ is projected to image space by $V=X_c[0:2,:]/Z$, where $Z= X_c[2,:]$ represents the distance from each vertex to camera. This illustrates that the distance $Z$ mainly leads to the difference of the two projections.

\begin{figure*}[t]
\centering
\includegraphics[height=7.7cm]{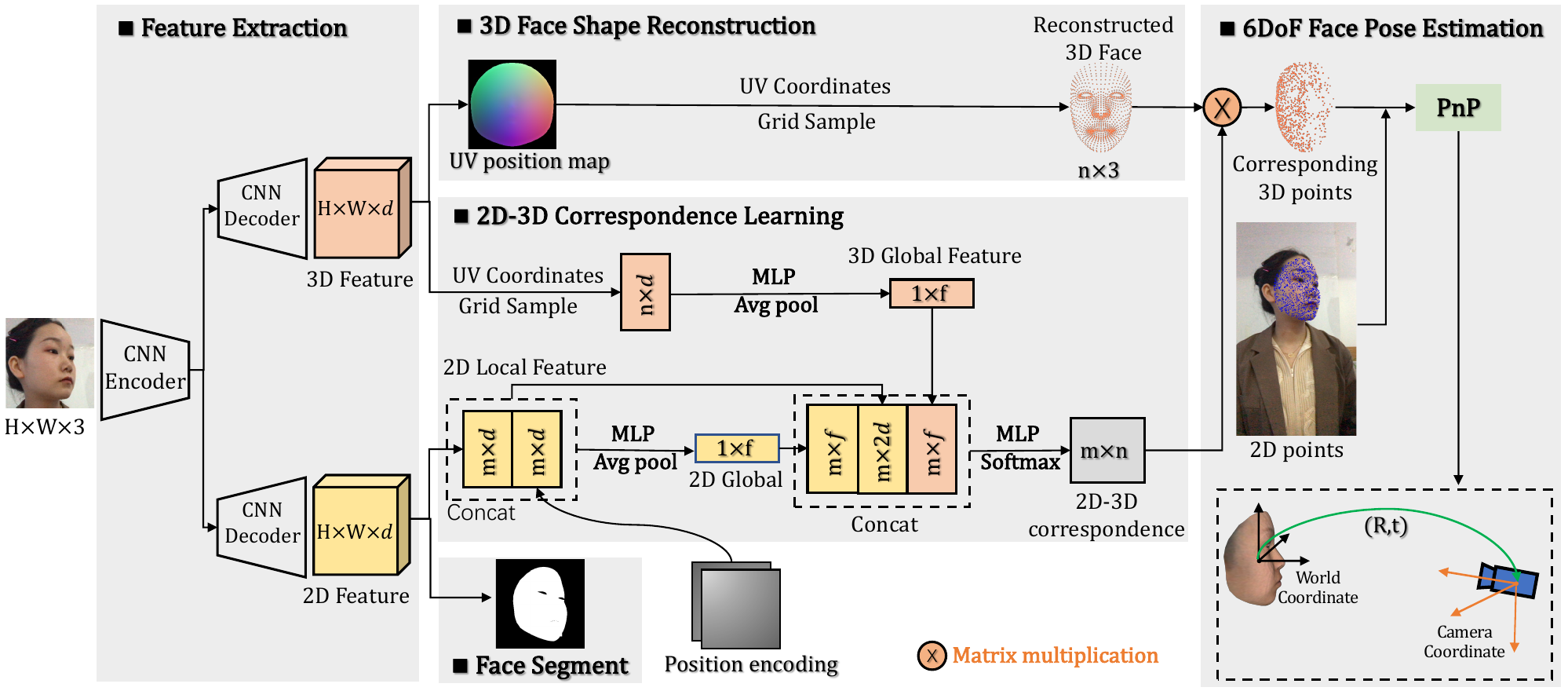}
\caption{The framework of our proposed method based on PerspNet. 3D features and 2D image features are extracted from encoder-decoder architectures respectively from a 2D facial image. The 3D features are fed into 3D face shape reconstruction module to predict 3D face shape information in world coordinate system. The  2D facial image features, 2D position encoding features and 3D features are fused learn the correspondence between 2D pixels and 3D points in reconstructed 3D face shape. With the corresponding 2D pixels and 3D points in a face, the 6DoF pose of the face can be computed by a PnP algorithm. In addition, 2D image features are also fed into 2D face segmentation, which is used to extract 2D pixels in face regions when testing.}
\label{fig:framework}
\end{figure*}

In this work, we focus on 3D face reconstruction under perspective projection, especially in very near distance. Given an RGB image $I \in \mathbb{R}^{H \times W \times 3}$, the goal is to find a method $F$ to recover the 3D face shape $X$ and estimate its 6DoF pose $(R,t)$: $ [X, R, t]  =F(I)$. It is achieved by a framework based on a new deep learning network, PerspNet, as shown in Fig.~\ref{fig:framework}. The proposed method contains two sub-tasks: 3D face shape reconstruction and 6DoF face pose estimation. For the 3D face shape reconstruction, we design a UV position map~\cite{feng2018joint} regression method. For 6DoF face pose estimation, we use a two-stage pipeline~\cite{peng2019pvnet,he2020pvn3d,tian2020shape}, that first chooses $m$ 2D pixels $V \in \mathbb{R}^{2 \times m}$ in 2D image and learns their corresponding 3D points in reconstructed 3D face shape and 6D face pose parameters can be calculated by a PnP~\cite{lepetit2009epnp} algorithm.

\subsection{Perspective Network (PerspNet)}
Specifically, PerspNet consists of four modules: feature extraction, 3D face shape $X$ reconstruction, 2D-3D correspondence $M$ learning and 2D face segmentation $c$. Given a 2D facial image $I$, PerspNet $G$ predicts $ [X, M, c]  = G(I)$. With the input, an encoder and two decoders are trained to extract 3D features $f_{3D}$ and 2D image features $f_{2D}$ respectively. The 3D features $f_{3D}$ are fed into 3D face shape reconstruction module to regress the UV position map, which represents the 3D face shape information in canonical space. Then the 2D features $f_{2D}$ includes 2D facial image features $f_{im}$ and encoded 2D position features $f_{pos}$,  and then fuse with 3D features $f_{3D}$ to learn the correspondence between 2D pixels in 2D image and 3D points in 3D face shape. With the correspondence, the 6DoF pose of the face can be computed. In addition, 2D image features are also fed into 2D face segmentation module, which is used to extract 2D observed pixels in face regions during testing.

\noindent
\textbf{3D Face Shape Reconstruction.}  Different from the UV formulation in \cite{feng2018joint} which is defined in image coordinate system, our UV position map $S \in \mathbb{R}^{H \times W \times 3}$ records the 3D coordinates of 3D facial structure with the canonical pose, which only represents the facial shape. Specifically, 
\begin{equation}
 S= Render(S_{coord}, X, T),
  \label{eq:first}
\end{equation}
where S is the rendered UV position map, $S_{coord} \in \mathbb{R}^{2 \times n}$ represents UV coordinates recording the 2D locations of $n$ 3D vertices $X$ in UV map, and $T \in \mathbb{R}^{k \times 3}$ is denoted as $k$ triangles in a 3D face mesh. A fully convolutional encoder-decoder architecture is utilized to regress the UV position map, as shown in Fig.~\ref{fig:framework}. 
To supervise the 3D face shape prediction, a weighted L1 loss function $L_{S}$ is used to measure the difference between ground-truth position map $S$ and the network output $\hat{S}$, 
\begin{equation}
 L_{S}=\sum ||S-\hat{S} || \cdot W(S_{coord}),
 \label{eq:first}
\end{equation}
 where $W \in \mathbb{R}^{H \times W}$ is a weight matrix for $S$, and we set $W(S_{coord})=1$, others in $W$ to 0. Then we extract the 3D vertices $X$ of the face from UV position map $S$ using UV coordinates.

\noindent
\textbf{2D-3D Correspondence Learning.} To estimate the 6DoF pose of the face in 2D images, we use a two-stage pipeline that first learns the correspondence of  2D pixels in 2D image and 3D points in reconstructed 3D face shape, and then compute 6D face pose parameters using a PnP algorithm. We design a 2D-3D correspondence learning module in PerspNet for the first stage, as shown in Fig.~\ref{fig:framework}. We build a correspondence probability matrix $M \in \mathbb{R}^{m \times n}$ where each row records the corresponding 3D points in canonical space of one pixel on the input image. Here $m$ is the number of face region pixels selected from a 2D image and $n$ is the number of vertices in 3D face shape. The estimating of M is as: 
\begin{equation}
 M=\text{softmax}(MLP(f_{2D}, f_{3D})).
 \label{eq:first}
\end{equation}
To learn the correspondence $M$ between 2D and 3D points, we extract 2D features $f_{2D}$ and 3D features $f_{3D}$ respectively. 

For 2D features $f_{2D}$, we firstly extract the image features $f_{im}$ from another fully convolutional decoder architecture after the encoder network from the same facial image. For each pixel in $V$, we also extract 2D position features $f_{pos}$ in 2D images and fuse $f_{im}$ as its local feature $f_{2Dlocal}$. Since the input image is a 2D spatial structure, the position features are encoded by a 2D position encoding method \cite{wang2019translating}, which is an extension of 1D position encoding from \cite{vaswani2017attention}. 2D global features $f_{2Dglobal}$ are also learned by feeding 2D local features $f_{2Dlocal}$ into Multi-Layer Perceptron (MLP) layers, followed with a global average pooling layer. Then the 2D local $f_{2Dlocal}$ and global features $f_{2Dglobal}$  are fused as 2D features $f_{2D}$.

As for the 3D  features $f_{3D}$, since the UV position map contains 3D geometry information, we also extract 3D global features $f_{3D}$ from the UV position regression network followed by MLP layers and a global average pooling layer. Then the 2D features $f_{2D}$ and 3D features $f_{3D}$ are fused and fed into MLP layers and a Softmax layer. In this way, the correspondence matrix $M$ is predicted from the network.

\begin{figure}[t]
\centering
\includegraphics[height=4.7cm]{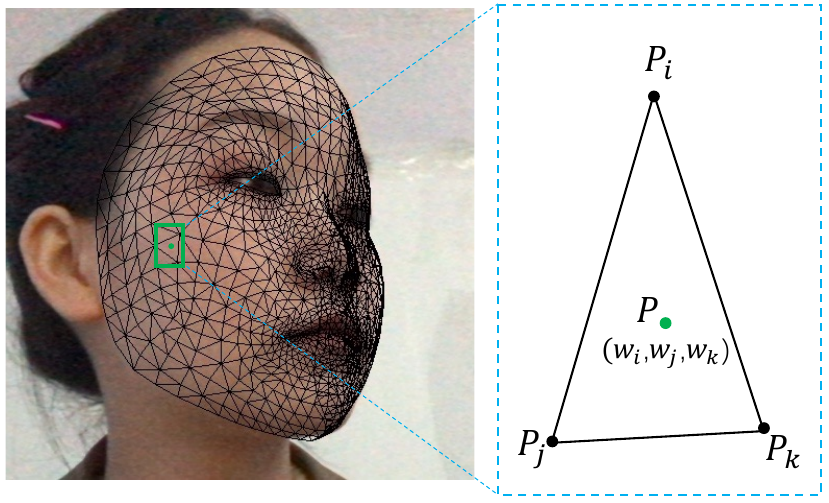}
\caption{A cropped 2D facial image with its projected  3D face triangles mesh in black. The barycentric coordinate of a pixel $P$ (green) in 2D facial image is calculated by three projected vertices $P_i, P_j, P_k$ of a triangle. } 
\label{fig:bc}
\end{figure}

To achieve the ground-truth $M$, for each pixel in the 2D face region, we compute its barycentric coordinates~\cite{weisstein2003barycentric} based on three vertices of its located triangle. The barycentric coordinates can be taken as corresponding probability between this 2D pixel and the three vertices in the 3D face mesh. Other values in $M$ are set to 0.  In this work, our 3D face mesh is a triangle mesh with $n$ 3D vertices and $k$ triangles. Each triangle consists of 3 vertices and 3 edges.  When a 3D face mesh is projected to a 2D image, a pixel $P$ in the 2D face region only belongs to a triangle with three vertices, $P_i,P_j, P_k$, as shown in Fig.~\ref{fig:bc}.  Its barycentric coordinate $(w_i,w_j,w_k)$ in this triangle, with one additional condition $w_i+w_j+w_k=1$, can be taken as corresponding probability between these 2D pixels and the three vertices in the 3D face mesh. Each row $M_i$ of the matrix $M$ represents the distribution over the correspondences between $i$-th pixels in $V$ and the vertices $X$. We utilize a Kullback-Leibler (KL) divergence loss for the correspondence matrix $M$. In addition, since the matrix $M$ is very sparse, we also minimize its entropy to regularize the matrix. The final loss for $M$ is   
\begin{equation}
 L_{M}= \frac{1}{m} (\sum_{i=0}^m D_{KL}(\hat{M_i}||M_i) - \lambda \sum_{i=0}^m \sum_{j=0}^n \hat{M}_{ij}\log \hat{M}_{ij} ).
  \label{eq:first}
\end{equation}
Here $M$ and $\hat{M}$ are ground-truth and predicted correspondence matrix, and $\lambda$ is a constant weight. With the predicted matrix $\hat{M}$ and reconstructed 3D face $\hat{X}$, the corresponding 3D points for each 2D pixels $X_{corr}$ are obtained by matrix multiplication, $\hat{X}_{corr}=\hat{M} \times \hat{X}^\mathrm{T}, X_{corr} \in \mathbb{R}^{m\times 3}.$
L1 loss $L_{corr}$ is also applied to supervise the $\hat{X}_{corr}$: 
\begin{equation}
L_{corr}=||X_{corr}-\hat{X}_{corr} ||_{L1}
  \label{eq:first}
\end{equation}
where ground-truth $X_{corr}$ is achieved by perspective projection.

\noindent 
\textbf{2D Face Segmentation.} 2D face segmentation module in the proposed network is trained for selecting 2D pixels from face regions in the inference phase. When training the whole network, the $m$ image pixels are randomly chosen from ground-truth face segmentation mask. While in the testing phase, the $m$ image pixels are randomly chosen from predicted face mask. 2D face segmentation task follows the 2D image feature extraction network and a 2-class softmax loss $L_{seg}$ is used. 

\noindent
\textbf{Training Objective.} We train out whole network with a multi-task loss. The final loss function is 
\begin{equation}
  L= \lambda_1 L_{S}+ \lambda_2 L_{M} +\lambda_3 L_{corr}+\lambda_4 L_{seg}.
  \label{eq:first}
\end{equation}
where $\lambda_1, \lambda_2, \lambda_3$ and $\lambda_4$ are the weights of the four losses respectively. Experimental results reveal that jointly training these tasks boosts the performance of each other.

\subsection{6DoF Face Pose Estimation}
Based on the output of the network, the final 6DoF pose estimation can be computed. Given the chosen 2D pixel coordinates $V$ in the original full 2D image, their corresponding 3D points coordinates $X_{corr}$ from reconstructed faces in world coordinate and the camera intrinsic parameters $K$, we apply a PnP algorithm~\cite{lepetit2009epnp} with Random Sample Consensus (RANSAC)~\cite{fischler1981random} to compute the 6D face pose parameters, $(R, t)=PnP(K, V, X_{corr})$. Perspective-n-Point is the problem of estimating the pose of a calibrated camera given a set of $m$ 3D points in the world and their corresponding 2D projections in the image. The camera pose consists of 6DoF which are made up of the rotation (roll, pitch, and yaw) and 3D translation of the camera with respect to the world. With the estimated 6DoF face pose, the reconstructed 3D face shapes can be projected to 2D images. It is worth noting that, directly regressing 6DoF pose parameters from a single image by CNN is also feasible, but it achieves much worse performance than our method due to the nonlinearity of the rotation space~\cite{peng2019pvnet}. It will be validated in our experiments.

\begin{table*}[t]
  \caption{Comparing ARKitFace with other 3D Face datasets. Exp. and Vert. are abbreviations of the annotation number of categories of expressions and number of vertices for 3D mesh, respectively.}
  \label{tab:dataset}
  \centering
   \setlength\tabcolsep{7pt}
  \begin{tabular}{l  c  c c c c c c c c c}
        \hline
    Dataset                                                                       & Sub. Num  &Image Num & 3D Mesh Num & Exp. Num   & camera & Vert. Num & 6DoF Pose\\
   
       \hline
    Bosphorus\cite{savran2008bosphorus}                    &    105       &  4,666         & 4,666                & 35              &Mega     &35K           & No \\
    BFM~\cite{paysan20093d}                                         &    200        & synthetic   & 200                  & Neutral       &ABW-3D &50K          & No \\
    FaceWarehouse~\cite{cao2013facewarehouse}        &    150        & 3000         & 3DMM              & 20               &Kinect v1 &20K         & No\\
    MICC\cite{bagdanov2011florence}                           &    53          & 53             & 203                  & $\leq$  5                &3DMD     &40K         & No\\
    3DFAW\cite{pillai20192nd}                                      &    26           & 26             & 26                   & Neutral           &DI3D      &20K         & No \\
    BP4D\cite{zhang2014bp4d}                                    &   41            & 328           &328                  &8                      & 3DMD    &70k        & No \\
    AFLW2000-3D\cite{zhu2017face}                                                &   2,000      & 2,000           &3DMM              &-                     & -    & 53,149        & No\\
    BIWI\cite{fanelli2013random}                                  &   20            & 15,678    &24                       &Neutral           & Kinect    &6,918        & Yes \\
    NoW\cite{sanyal2019learning}                                 &   100            & 2,054           &100              &Neutral         & iPhone X   &58,668        & No \\
    FaceScape\cite{yang2020facescape}                      &   938     &   1,275,680         &18,760                  &20                      & DSLR    &2M        & fixed pose \\

       \hline
    ARKitFace                                                              &   500            & 902,724          & 902,724                 &33                     & iPhone 11    &1,220        & Yes \\
         \hline
  \end{tabular}

\end{table*}

\begin{figure*}[t]
\centering
\includegraphics[height=6.7cm]{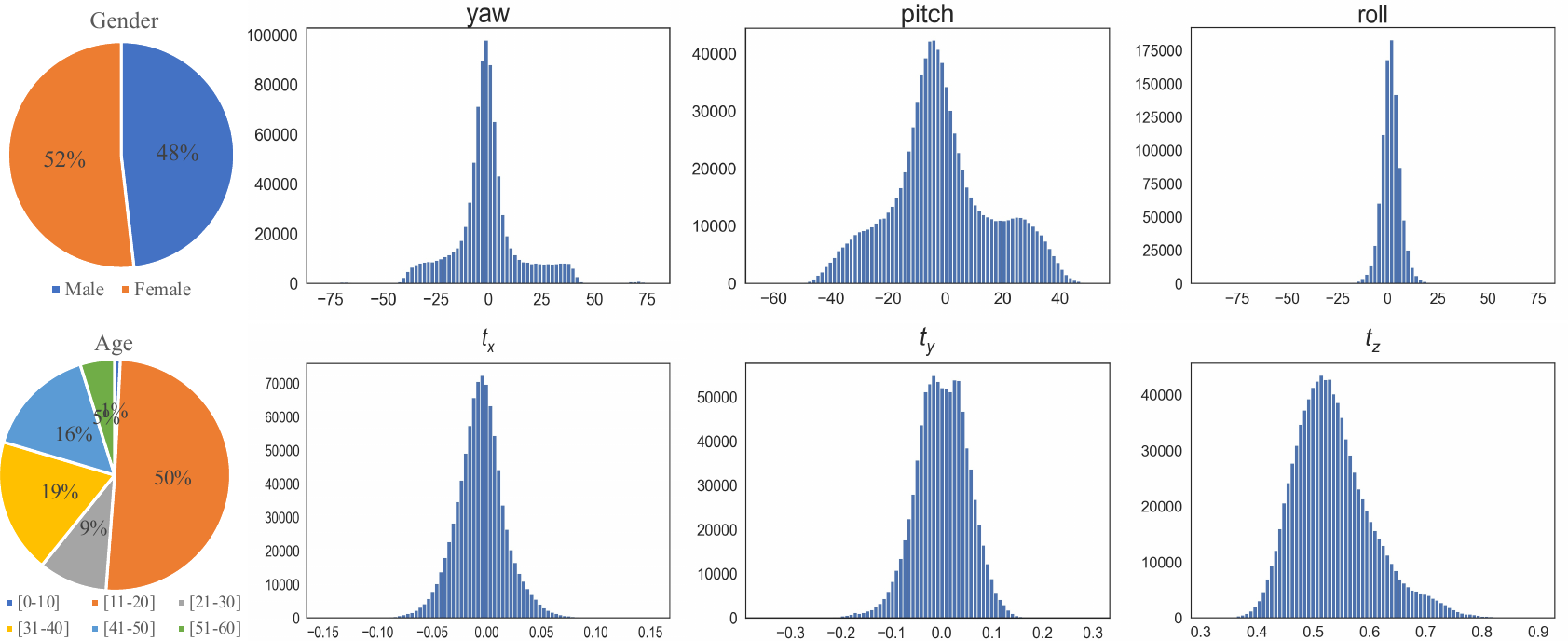}
\caption{Distributions of age, gender, and each pose parameter of 6DoF on ARKitFace dataset.} 
\label{fig:dataset}
\end{figure*}

\section{ARKitFace Dataset} 
The ARKitFace dataset is established by this work in order to train and evaluate both 3D face shape and 6DoF in the setting of perspective projection. A total of 500 volunteers, aged 9 to 60, are invited to record the dataset. They sit in a random environment, and the 3D acquisition equipment is fixed in front of them, with a distance ranging from about 0.3m to 0.9m. Each subject is asked to perform 33 specific expressions with two head movements (from looking left to looking right / from looking up to looking down). 3D acquisition equipment we used is an iPhone 11. The shape and location of human face are tracked by structured light sensor. The triangle mesh and 6DoF information of the RGB images are obtained by built-in ARKit toolbox. The triangle mesh is made up of 1,220 vertices and 2,304 triangles. In total, 902,724 2D facial images (resolution $1280 \times 720$  or $1440 \times 1280$) with ground-truth 3D mesh and 6DoF pose annotation are collected. An example is shown in Fig.~\ref{fig:arkit_face_example}. 

Distributions of age, gender, and each pose parameter of 6DoF on ARKitFace dataset are shown in Fig.~\ref{fig:dataset}. We can observe that our dataset has balanced gender, diverse age and the 6DoF pose variation. Comparisons between different datasets shown in Table \ref{tab:dataset} reveal that ARKitFace surpasses the existing datasets in terms of scale, 3D exact shape annotations and diversity of poses.
 In our ARKitFace dataset, the yaw angle ranges from -$78^{\circ}$ to $77^{\circ}$, pitch angle ranges from -$64^{\circ}$ to $51^{\circ}$ and roll angle ranges from -$90^{\circ}$ to $63^{\circ}$. The examples with large poses can be seen in Fig.~\ref{fig:arkit_face_example} (d) and Section~\ref{sec:qua}. This distribution of pose parameters is enough for most of real-world applications, especially with selfies, such as virtual glasses try-on and makeup.

\noindent 
\textbf{Authorization:} All the 500 subjects consent to use their data. We will release all the subjects with 2D facial images, 3D mesh and 6DoF pose annotation under the authorization of all the subjects. We also prepare the model parameters of our 3D model for further utilization. We will not release their personal privacy information, including age, gender etc. We have uploaded the data of a subject to IEEE DataPort (DOI:10.21227/jfbk-0j17, Title: A Subject Data of ARKitFace Dataset) and take it as an example.

\section{Experiments}
\subsection{Implementation Details}
Our PerspNet is implemented by PyTorch~\cite{paszke2019pytorch}. During training, the PerspNet takes the input image cropped from a full 2D image and resized to $192 \times 192 \times 3$, based on the ground-truth face segmentation mask. To augment the data with large poses, we utilize face profiling method~\cite{zhu2016face} to generate the profile view of faces from medium-pose samples for three euler angles. We enlarge all the three angles to max $90^{\circ}$ and min -$90^{\circ}$. We also apply online data augmentation including random cropping, resizing and color jittering during training. We use a pre-trained ResNet-18~\cite{he2016deep} architecture, where the final encoded feature map is $6 \times 6 \times 512$. To regress the $192 \times 192 \times 3$ UV position map, the first decoder is implemented by 5 up-sampling layers and its output is a $192 \times 192 \times 32$ feature map, which is regarded as the 3D features. The number of point clouds $n$ in 3D face shape is 1220. To extract 2D image features and segment 2D face region from background, the second decoder consists of five up-sampling layers and each up-sampled feature map is concatenated with a feature map which have the same size in the encoder backbone network. The size of 2D image features is also $192 \times 192 \times 32$. The number of randomly sampled 2D pixels $m$ for face region is 1024. If point count in face region is insufficient, we sample these pixels by repetition. Since the size of input image is 192x192, the point count in the face region is sufficient for 1024 points in most cases. The repeated sampling is few and side effect can be ignored. The 2D and 3D global feature sizes are all $1 \times 1024.$ At the training phase, the 2D pixels are randomly chosen from ground-truth face mask. We set the weights of losses respectively, $\lambda_1=0.5, \lambda_2=0.01, \lambda_3=1.0, \lambda_4=0.01, \lambda=0.1$. We train all models for 20 epochs. The first 10 epochs with initial learning rate, which is set as 0.0001. Then the learning rate is linearly decayed to zero over the left 10 epochs. All the networks are trained and evaluated on the ARKitFace dataset with ground-truth bounding box. At the testing phase, the 2D pixels are randomly chosen from segmented face region. The PnP algorithm is implemented in OpenCV~\cite{bradski2000opencv}.

\subsection{Dataset}
To validate our proposed method, we conduct experiments on our collected ARKitFace dataset and a public BIWI dataset.

\noindent 
\textbf{ARKitFace.} We randomly use 400 people in our dataset as training data with a total of 717,840 2D facial images and annotations, leaving 100 people with totally 184,884 samples for testing.

\begin{table*}[t]
\caption{Comparisons with different methods for 6DoF Face Pose Estimation on ARKitFace test dataset.(bbox: bounding box, pred: detected bbox, GT: GT bbox. mask: facial mask, pred: segment mask, GT: GT mask.)}
  \label{tab:pose}
  \centering
   \setlength\tabcolsep{6pt}
  \begin{tabular}{l  c c c c c c c c c c c}
         \hline
    Method & bbox     &mask      &Yaw $\downarrow$ & Pitch $\downarrow$ & Roll $\downarrow$ & $MAE_r \downarrow$ & $t_x \downarrow$ & $t_y \downarrow$ &$t_z \downarrow$ & $MAE_t \downarrow$ & ADD $\downarrow$\\
   \hline
    img2pose(retrain)~\cite{albiero2021img2pose}& pred & pred   &5.07&7.32&4.25&5.55&1.39&3.72&15.95&7.02&20.54 \\
    Direct 6DoF Regress   & --        & pred                           & 1.86& 2.72 & 1.03 &1.87   &2.80 &5.23 &19.16 &  9.06    & 21.39 \\
    Refined Pix2Pose\cite{park2019pix2pose}   & pred     & pred    & 1.95 & 2.62  & 2.48  & 2.35  & 2.43  & 4.23  &35.33 &  14.00     & 36.44  \\ 
    PerspNet w/o PE & pred     & pred                         & 1.00 & 1.52 & 0.62 & 1.05 &1.18 &2.39 &12.14 &  5.24    & 12.71 \\ 
    PerspNet w/o $L_{M}$  &pred       & pred            & 1.04 & 1.45 & 0.59 & 1.03  &1.14 & \textbf{2.21} &10.84 &  4.73    & 11.45\\
    PerspNet (ours) &pred           & pred                & \textbf{ 0.98} &  \textbf{1.43} &  \textbf{0.55} &   \textbf{0.99} & \textbf{1.00} & 2.41 & \textbf{9.73} &   \textbf{4.38}    &  \textbf{10.30} \\ 
     \hline
     \hline
    PerspNet w/o PE         &GT       & pred               & 1.01 & 1.53 & 0.61 & 1.05  &1.17 &2.39 &11.77 &  5.11    & 12.34 \\    
    PerspNet w/o $L_{M}$  &GT    & pred                & 1.04 & 1.45 & 0.60 & 1.03  &1.09 &2.13 &10.28 &  4.50    & 10.89\\
    PerspNet (ours)      &GT       & pred        & \textbf{0.99} & \textbf{1.43} & \textbf{0.55} & \textbf{0.99}  &\textbf{0.97} &\textbf{2.12} &\textbf{9.45} &   \textbf{4.18}   & \textbf{10.01} \\
        \hline
          \hline
    PerspNet w/o PE  &    GT       &    GT             & 0.94 & 1.38 & 0.59 & 0.97  &0.94 &1.97 &10.64 &  4.52    & 11.07\\
    PerspNet w/o $L_{M}$   &   GT   &    GT         &0.78  &1.15  &  \textbf{0.54}&  0.82 & 0.99& 1.58&9.88  &    4.15   &10.34 \\
    PerspNet (ours) &   GT        &    GT            & \textbf{0.72} & \textbf{1.10} & \textbf{0.54} & \textbf{0.79}  & \textbf{0.92} & \textbf{1.47} & \textbf{9.59} &   \textbf{ 3.99}   & \textbf{9.99} \\ 
     \hline
  \end{tabular}
\end{table*}

\begin{table*}[t]
 \caption{Comparisons with different methods for 6DoF Face Pose Estimation on BIWI dataset.}
  \label{tab:posebiwi}
  \centering
  \setlength\tabcolsep{6pt}
  \begin{tabular}{l  c  c c c c c c c c c}
         \hline
    Method                               &Yaw $\downarrow$ & Pitch $\downarrow$ & Roll  $\downarrow$ & $MAE_r \downarrow$ & $t_x \downarrow$ & $t_y \downarrow$ &$t_z \downarrow$ & $MAE_t \downarrow$ & ADD $\downarrow$\\
    \hline
    Dlib (68 points)\cite{kazemi2014one}                  & 16.76 & 13.80 & 6.19 &  12.25& -& -&-&-&-\\
    3DDFA\cite{zhu2016face}                               & 36.18 & 12.25 & 8.78 &  19.07&  -& -&-&-&-\\
    FAN (12 points)\cite{bulat2017far}                   & 8.53 & 7.48 & 7.63 &  7.88&  -& -&-&-&-\\
    Hopenet ($\alpha =1$)\cite{ruiz2018fine}        & 4.81 & 6.61 & 3.27 &  4.90&  -& -&-&-&-\\
    QuatNet\cite{hsu2018quatnet}                               & 4.01 & 5.49 & 2.94 &  4.15&  -& -&-&-&-\\
    FSA-NET\cite{yang2019fsa}                             & 4.56 & 5.21 & 3.07 &  4.28&  -& -&-&-&-\\
    HPE\cite{huang2020improving}                                     & 4.57 & 5.18 & 3.12 &  4.29&  -& -&-&-&-\\
    TriNet\cite{cao2021vector}                       & \textbf{3.05} & 4.76 & 4.11 &  3.97&  -& -&-&-&-\\

    RetinaFace R-50(5 points)\cite{deng2020retinaface}  & 4.07 & 6.42 & 2.97 &  4.49&  -& -&-&-&-\\
    img2pose\cite{albiero2021img2pose}                    & 4.57 & 3.55 & 3.24 &  3.79 &  -& -&-&-&- \\

     Direct 6DoF Regress                   &16.49  & 14.03 & 5.81 &  12.11 &62.36  &85.01  &366.52  & 171.30 &562.38  \\ 
      Refined Pix2Pose\cite{park2019pix2pose}   & 5.75 & 5.06  & 11.23  & 7.35 &16.82 &21.30  &255.36 &  97.83     & 356.32 \\ 
     PerspNet w/o PE                         &3.63  & 3.81 & 3.48 & 3.64  &6.03  &9.11  &77.87  & 31.00 &142.20  \\ 
    PerspNet w/o $L_{M}$                 &3.67  & 3.52 & 3.26 & 3.48 &5.57  &8.53  &75.23  &29.78  &136.16  \\ 
    PerspNet (ours)               &  3.10 &  \textbf{3.37} &  \textbf{2.38} &  \textbf{2.95}  &  \textbf{4.15}& \textbf{6.43}&\textbf{46.69}&\textbf{19.09}&\textbf{100.09} \\
   
       \hline
  \end{tabular}
\end{table*}

\noindent 
\textbf{BIWI.} BIWI~\cite{fanelli2013random} contains 24 videos of 20 subjects in an indoor environment. Each video is coupled with a neutral 3D face mesh of a specific person. There are totally 15, 678 frames with a wide range of face poses in this dataset. Paired RGB and Depth images are provided for each frame. We only use the RGB images as inputs in this work. This benchmark provides ground-truth labels for rotation (rotation matrix) and translation for full 6DoF. Since there is not each 3D face mesh for each frame, we can not evaluate our method on 3D face reconstruction task. We only evaluate our method for 6DoF pose estimation task. In addition, we train our method on training set of ARKitFace dataset, and test our method on the entire BIWI dataset following the previous methods~\cite{hsu2018quatnet,yang2019fsa,albiero2021img2pose}.

\subsection{Evaluation Metric}
For the 3D face shape reconstruction, we follow previous works~\cite{feng2021learning,sanyal2019learning}, median distance and average distance between predicted 3D mesh vertices and ground-truth 3D mesh vertices are utilized. For the 6DoF face pose estimation, we follow previous head/face pose estimation methods~\cite{albiero2021img2pose,yang2019fsa,hsu2018quatnet}, and convert the rotation matrixes to 3 euler angles, Yaw, Pitch, Roll  and compute the mean absolute error (MAE) for 6DoF, $Yaw, Pitch, Roll, t_x,t_y,t_z $. $MAE_r$ and $MAE_t$, the rotational and translational MAE are also computed. Furthermore, to validate the 6DoF face pose in a metric, we adopt the average 3D distance (ADD) metric~\cite{hinterstoisser2012model} used for object pose evaluation. Given the ground-truth rotation $R$ and translation $t$ and the estimated rotation $\hat{R}$ and translation $\hat{t}$, the ADD computes the mean of the pairwise distances between the ground-truth 3D face model points $x \in X$ transformed based on the ground-truth pose and the estimated pose: $ADD= \text{avg}_{x \in X} ||(Rx+t) - (\hat{R}x+\hat{t} )||$.

 \begin{figure*}[t]
\centering
\includegraphics[height=13.3cm]{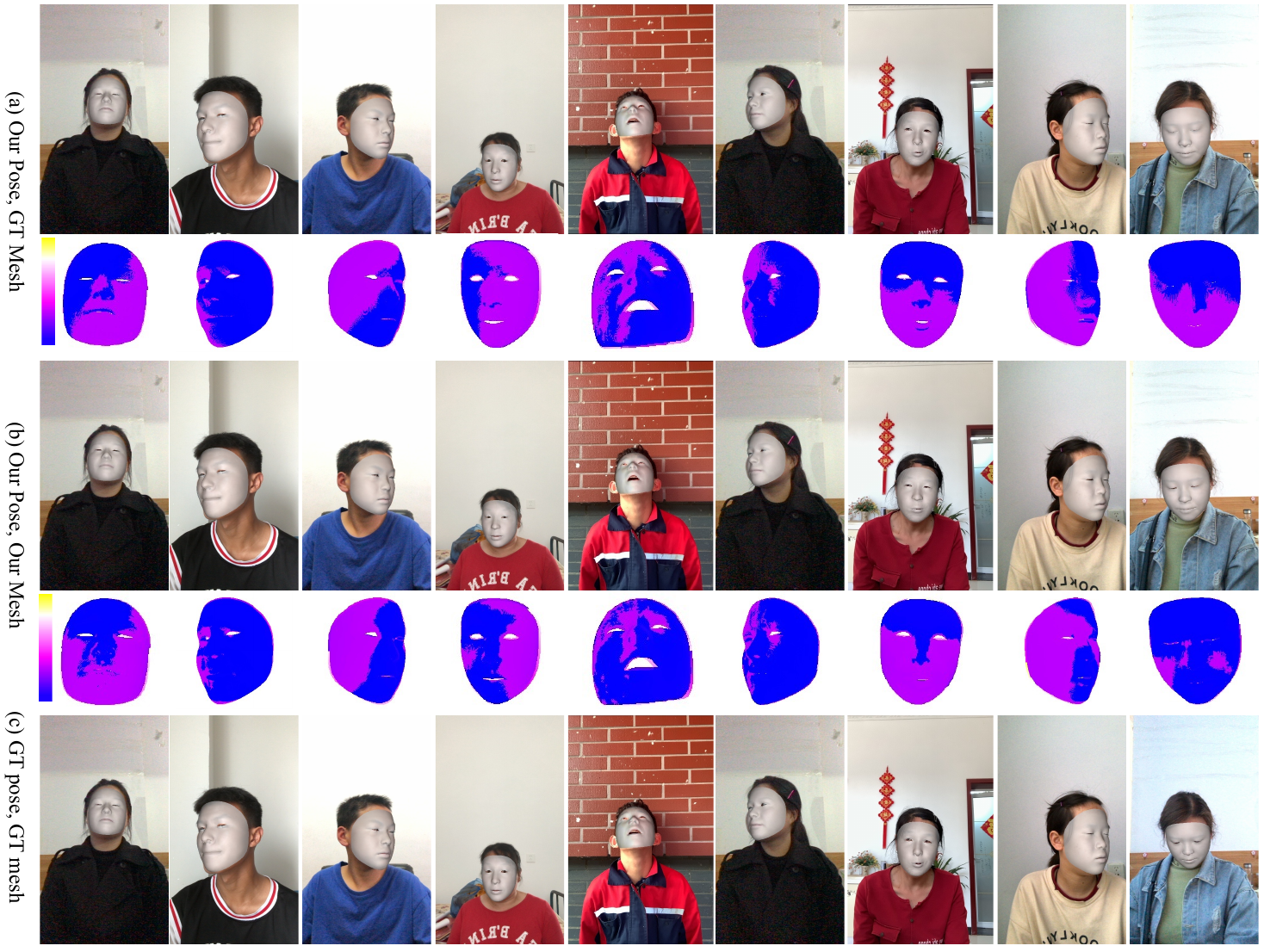}
\caption{Qualitative results for 6DoF pose estimation and 3D face shape reconstruction on ARKitFace dataset.}
\label{fig:ar}
\end{figure*}

\begin{table*}[t]
 \caption{Ablation Studies for 6DoF Face Pose Estimation on entire BIWI dataset (15, 678 images) with GT bounding box.}
  \label{tab:biwi_gt}
  \centering
  \setlength\tabcolsep{6pt}
  \begin{tabular}{l  c  c c c c c c c c c}
         \hline
    Method                               &Yaw $\downarrow$ & Pitch $\downarrow$ & Roll $\downarrow$ & $MAE_r \downarrow$ & $t_x \downarrow$ & $t_y \downarrow$ &$t_z \downarrow$ & $MAE_t \downarrow$ & ADD $\downarrow$\\
    \hline
     PerspNet w/o PE                       &4.41  & 5.32 & 6.13 & 5.29  & 6.97 &12.61  &101.62 & 40.40 &187.81  \\ 
     PerspNet w/o $L_{M}$                &4.40  & 4.52 & 5.08 & 4.67 &6.65  &11.36  &87.29  & 35.10 &162.97  \\ 
     PerspNet (ours)               &   \textbf{3.44} &  \textbf{4.39} &  \textbf{3.09} &  \textbf{3.64}  &  \textbf{4.21}& \textbf{8.02}&\textbf{52.27}&\textbf{21.50}&\textbf{113.53} \\     
       \hline
  \end{tabular}
\end{table*}

\begin{figure*}
\centering
\includegraphics[height=18.65cm]{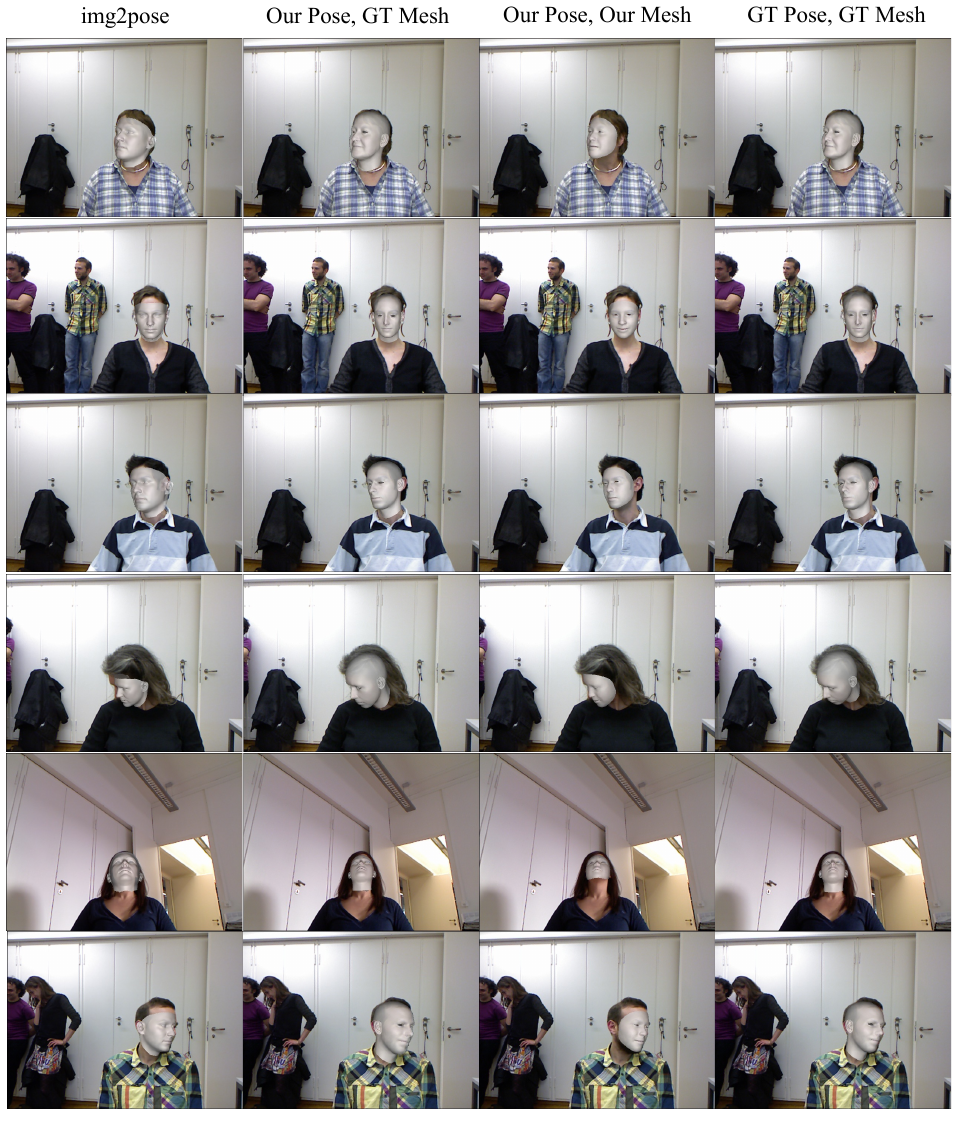}
\caption{Qualitative results for 6DoF pose estimation and 3D face shape reconstruction on BIWI dataset.}
\label{fig:biwir}
\end{figure*}

\subsection{Evaluation for 6DoF Face Pose Estimation}
\noindent
\textbf{Comparison with the state-of-the-art methods.}  To compare with the state-of-the-art methods on head or face pose estimation, we firstly retrain the most recent state-of-the-art method, img2pose~\cite{albiero2021img2pose}, on ARKitFace training set, and compute the performance on testing data. As shown in Table~\ref{tab:pose}, our method outperforms it significantly. We use the code of \cite{bulat2017far} to detect the facial 68 landmarks and crop facial region. In addition, the public BIWI dataset is used as a cross-data evaluation to test our final network. Since some faces with large angles can not be detected, we follow the img2pose~\cite{albiero2021img2pose} method and test on 13,219 images with detected bbox. We also use the code of \cite{bulat2017far} to detect the face 68 landmarks and crop facial region. All the results are shown in Table~\ref{tab:posebiwi}. We can see that our method performs much better than previous methods, especially the recent img2pose method~\cite{albiero2021img2pose}. The experiment also demonstrates that our method and dataset can be well generalized to the data in different domain. Furthermore, we refine an object 6DoF pose estimation method, Pix2Pose\cite{park2019pix2pose}, to adapt to 6DoF face pose estimation. Specifically, Pix2Pose is a PnP-based method with the known 3D model of object. We use a mean face shape instead of the known 3D model of object and train it on ARKitFace training dataset. Its results on ARKitFace test dataset and BIWI dataset are shown in Table~\ref{tab:pose} and Table~\ref{tab:posebiwi}. We can see that our method that is specially designed for face has much better results. The inference time of our proposed model is 11.7ms with a P100 GPU.

\begin{figure*}
\centering
\includegraphics[height=6.9cm]{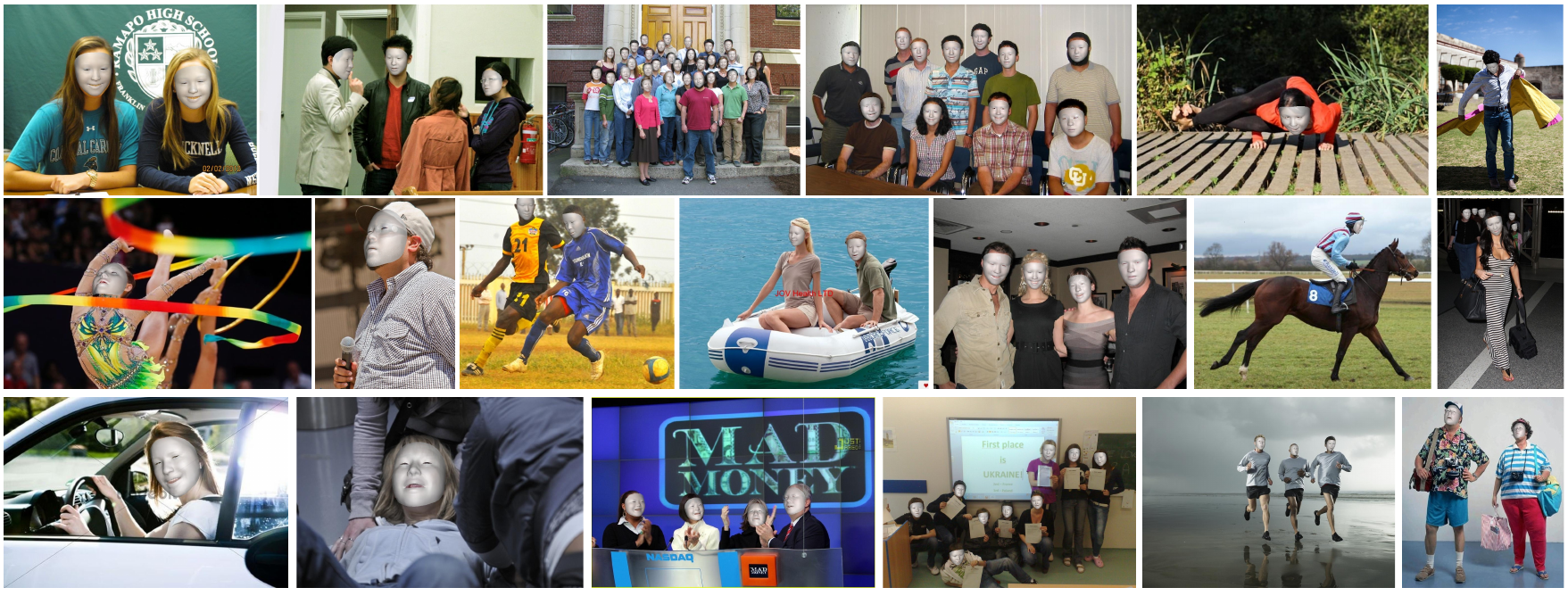}
\caption{Qualitative results for 6DoF pose estimation and 3D face shape reconstruction on WIDER FACE validation images.}
\label{fig:widerface}
\end{figure*}

\noindent
\textbf{Ablation Studies.} We build several baselines to evaluate the components that contribute to our performance. Since img2pose~\cite{albiero2021img2pose} directly regresses the 6 pose parameters, we build a baseline, Direct 6DoF Regress, to regress the 6 pose parameters after the backbone network.  Other baselines include our PerspNet without the position encoding features (PerspNet w/o PE), which is used to validate the effectiveness of the 2D position encoding features, and our PerspNet without $L_M$ loss (PerspNet w/o $L_M$),  which are built to evaluate the corresponding components. The results are shown in Table~\ref{tab:pose} and Table~\ref{tab:posebiwi}. We can observe that our two-stage method outperforms direct regression method significantly, and 2D position encoding features are helpful and $L_M$ is effective for face pose estimation task.

To explain the influence of detected bounding box, we provide the results with GT bounding box in Table \ref{tab:pose}. It shows that the method with the GT bounding box performs better than that with predicted bounding box, which indicates that more accurate detection results are needed. Moreover, to explain the influence of segmentation mask, the predicted and GT face masks are used during the inference time respectively. Its results are shown in Table \ref{tab:pose} respectively. It shows that the method with the ground-truth face mask performs better than that with predicted segmentation mask, which indicates that more accurate segmentation results are needed. In addition, since some left images except 13,219 images on BIWI dataset are failed to detect, we give the results with GT bounding box on all 15, 678 images, shown in Table \ref{tab:biwi_gt}. The results also prove that 2D position encoding features are helpful and $L_M$ is effective for face pose estimation task. We also see that the performance of 13,219 images in Table \ref{tab:posebiwi} is better than that on all dataset (15, 678 images) in Table \ref{tab:biwi_gt}. Because the left images except 13,219 images are hard samples with large angles for pose estimation.

\begin{table}
 \caption{Results for 3D face shape reconstruction on ARKitFace.}
  \label{tab:example}
  \centering
  \setlength\tabcolsep{5pt}
  \begin{tabular}{l  c c c }
        \hline
    Method & Median(mm) & Mean(mm) \\
         \hline
    PRNet~\cite{feng2018joint} &   1.97        &  2.05     \\
    3DDFA\_v2~\cite{guo2020towards} & 2.35                     &      2.31             \\
    Deng \emph{et al.}~\cite{deng2019accurate} &  2.46                   &      2.55            \\
    PerspNet (ours) &             \textbf{1.72}         &    \textbf{1.76}                 \\
       \hline
  \end{tabular}
 \end{table}

\subsection{Evaluation on 3D Face Shape Reconstruction}
To validate the proposed method on 3D face shape reconstruction task, we display the results on ARKitFace test data in Table~\ref{tab:example}. For comparison, we also train a single-task PRNet~\cite{feng2018joint} with the same encoder-decoder UV regression network in our multi-task network on ARKitFace training data. As shown in Table~\ref{tab:example}, our multi-task network outperforms the single-task PRNet, which reveals that the pose estimation task contributes to the improvement of 3D face shape reconstruction task. In addition, we compare our method with other SOTA like 3DDFA\_v2~\cite{guo2020towards},  Deng \emph{et al.}~\cite{deng2019accurate} in Table~\ref{tab:example}. We can see that our method still achieves the best performance of 3D face reconstruction.

\begin{figure}[t]
\centering
\includegraphics[width=1 \columnwidth]{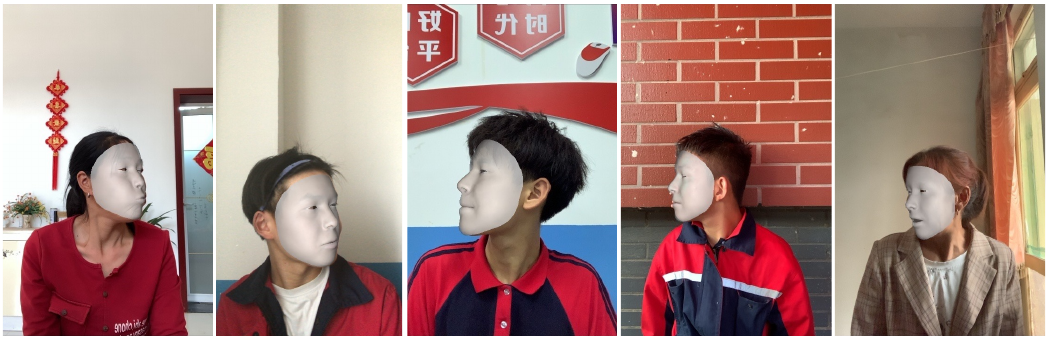}
\caption{Qualitative results with large poses for 6DoF face pose estimation on ARKitFace test dataset.}
\label{fig:largepose}
\end{figure}

\begin{figure}
\centering
\includegraphics[width=1 \columnwidth]{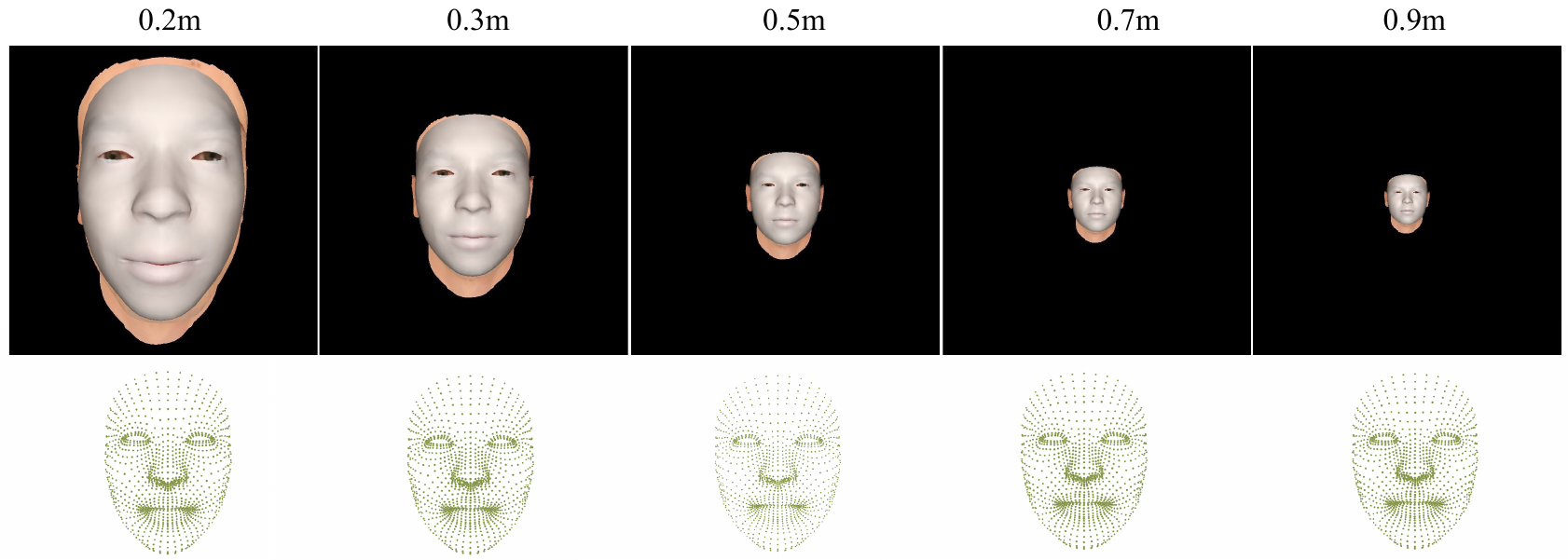}
\caption{Qualitative results on synthetic data with different distance from camera to 3D face center.}
\label{fig:bfm}
\end{figure}

\begin{figure}[t]
\centering
\includegraphics[width=0.9 \columnwidth]{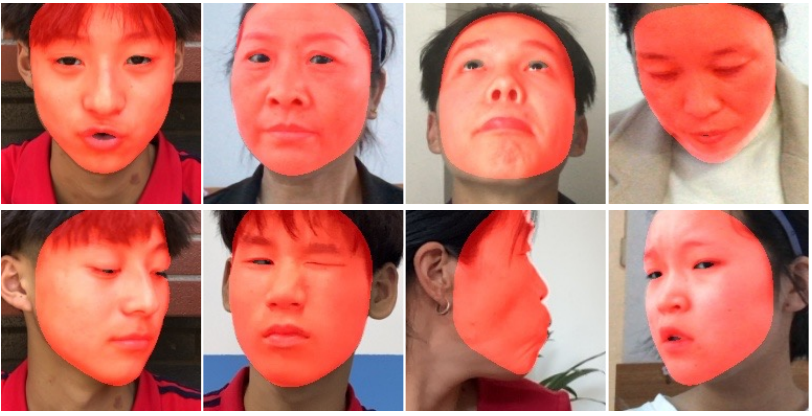}
\caption{Qualitative results for 2D face segmentation on ARKitFace test dataset.}
\label{fig:seg}
\end{figure}

\subsection{Qualitative Results for 6DoF Face Pose Estimation and 3D Face Shape Reconstruction}
\label{sec:qua}

We display qualitative results for 6DoF pose estimation and 3D reconstruction on ARKitFace and BIWI dataset in Fig.~\ref{fig:ar} and Fig.~\ref{fig:biwir} , where img2pose, our predicted results and GT results participate in the comparison. The predicted face pose with GT 3D mesh, the predicted face pose with predicted 3D mesh and their error map are demonstrated, respectively. The error map is the error between two kinds of meshes and GT mesh with GT pose, and is shown as a heat map in Fig.~\ref{fig:ar}. We can see that our method is effective and outperforms img2pose, especially in large pose. We also show some qualitative results on in-the-wild images from WIDER FACE dataset in Fig.~\ref{fig:widerface}. It shows that our method also performs well in in-the-wild images. To show the performance of our proposed method with large poses, we show some results with predicted face pose and GT 3D mesh on ARKitFace in Fig.~\ref{fig:largepose}, whose absolute yaw angles are all more than $45^{\circ}$. We can see that our method performs well as the yaw angle increases.

To further evaluate our method, we render synthetic data with different distance from camera to 3D face center (located about 8cm behind the nose) same with our ARKitFace dataset. Fig.~\ref{fig:bfm} shows the projection for a same subject from Basel Face Model (BFM)~\cite{paysan20093d} with our predicted 6DoF pose and  3D mesh with different $tz$. Each image is coupled with our predicted 3D vertices in canonical space. The results show that our method reconstructs consistent shapes regardless of the distances to the camera, showing robustness to the distortion brought by perspective projection.

\subsection{Evaluation on 2D Face Segmentation}
To validate the proposed method on 2D face segmentation task, we evaluate the segmented results in 3 metrics, Accuracy: 97.92\%, IOU (Intersection over Union): 95.50\%, Precision: 97.60\%. We also display qualitative results for 2D face segmentation on ARKitFace test dataset in Fig.~\ref{fig:seg}. We can see that our method performs very well.

\section{Conclusion}
We explore 3D face reconstruction under perspective projection from a single RGB image for 3D face AR applications. We introduce a novel framework, in which a deep learning network, PerspNet, is proposed, for 3D face shape reconstruction, corresponding learning between 2D pixels and 3D points in 3D face models, and 2D face region segmentation. With 2D pixels in facial images and corresponding 3D points in reconstructed 3D face mesh, 6DoF face pose is estimated by a PnP method. This 6DoF face pose is used for perspective projection transformation. To enable our PerspNet, we build a large-scale 3D face dataset, ARKitFace dataset, annotating 2D facial images, 3D face mesh and 6DoF pose. Experiments demonstrate the effectiveness of our approach for 3D face shape reconstruction and 6DoF pose estimation.  As most of the face analysis methods, our method and data may raise privacy concerns when misused. Therefore, the release of the data is fully authorized by the subjects. We wish this work would spur the future researches including 3D face reconstruction and face pose estimation.

{\small
\bibliographystyle{IEEEtran}
\bibliography{egbib}
}

\vfill

\end{document}